\documentclass{article} %
\usepackage[preprint]{colm2025_conference}

\usepackage{wrapfig}
\usepackage{microtype}
\usepackage{hyperref}
\usepackage{url}
\usepackage{tabularx}
\usepackage{booktabs}
\usepackage{graphicx}
\usepackage{subcaption}
\usepackage{amsmath}
\usepackage{amssymb}
\usepackage{fancyvrb, framed, inconsolata}
\usepackage[T1]{fontenc}

\usepackage{lineno}

\definecolor{darkblue}{rgb}{0, 0, 0.5}
\hypersetup{colorlinks=true, citecolor=darkblue, linkcolor=darkblue, urlcolor=darkblue}

\setcitestyle{maxnames=5}

\title{Using Scaling Laws for Data Source Utility Estimation in Domain-Specific Pre-Training}

\author{Oleksiy Ostapenko\textsuperscript{1}, Charles Guille-Escuret\textsuperscript{2}\thanks{Work done during internship at ServiceNow Research.}, Luke Kumar\textsuperscript{1}, Max Tian\textsuperscript{3}, Denis Kocetkov\textsuperscript{1}, \\ \textbf{Gopeshh Subbaraj\textsuperscript{2,$\ast$}, Raymond Li\textsuperscript{1}, Joel Lamy-Poirier\textsuperscript{1}, Sebastien Paquet\textsuperscript{1}, Torsten Scholak\textsuperscript{1}}  \\
\textsuperscript{1}ServiceNow Research \space\space\space 
\textsuperscript{2}Mila — Quebec AI Institute \space\space 
\textsuperscript{3}Reka AI \\
}

\begin{document}

\ifcolmsubmission
\linenumbers
\fi

\maketitle

\begin{abstract}

We introduce a framework for optimizing domain-specific dataset construction in foundation model training. Specifically, we seek a cost-efficient way to estimate the quality of data sources (e.g. synthetically generated or filtered web data, etc.) in order to make optimal decisions about resource allocation for data sourcing from these sources for the stage two pre-training phase, aka annealing, with the goal of specializing a generalist pre-trained model to specific domains. %
Our approach extends the usual point estimate approaches, aka micro-annealing, to estimating scaling laws by performing multiple annealing runs of varying compute spent on data curation and training. This addresses a key limitation in prior work, where reliance on point estimates for data scaling decisions can be misleading due to the lack of rank invariance across compute scales --- a phenomenon we confirm in our experiments. 
By systematically analyzing performance gains relative to acquisition costs, we find that scaling curves can be estimated for different data sources. Such scaling laws can inform cost effective resource allocation across different data acquisition methods (e.g. synthetic data), data sources (e.g. user/web data) and available compute resources. 
We validate our approach through experiments on a pre-trained model with 7 billion parameters. We adapt it to: a domain well-represented in the pre-training data --- the medical domain, and a domain underrepresented in the pretraining corpora --- the math domain. We show that one can efficiently estimate the scaling behaviors of a data source by running multiple annealing runs, which can lead to different conclusions, had one used point estimates using the usual micro-annealing technique instead. This enables data-driven decision-making for selecting and optimizing data sources.

\end{abstract}

\section{Introduction}
\label{introduction}

Large Language Models (LLMs)~\citep{brown2020language} have demonstrated remarkable versatility, acquiring a wide range of capabilities from pretraining on vast and diverse data corpora. However, in many real-world applications, generalist performance is not sufficient: there is an increasing need to specialize models for specific domains or tasks. One common strategy to address this is late-stage annealing, where domain-specific data is up-sampled and the learning rate is linearly annealed to zero~\citep{olmo20242, grattafiori2024llama3herdmodels, blakeney2024does}.  While this technique has shown promise in enhancing performance on targeted tasks, it remains unclear how to reliably estimate the utility of domain-specific data sources prior to large resource commitments.

A wide range of methods exists for acquiring domain-specific training data, each with distinct strengths, limitations, and cost structures~\citep{guo2022domain,olmo20242,cheng2023adapting}. Human annotation, while often considered the gold standard, is expensive and is mostly unfeasible to obtain at the pre-training scale. Model-based filtering (MBF) can efficiently extract high-relevance data from existing corpora, though it does not generate truly novel information beyond its source. Synthetic data generation leveraging other LLMs offers some degree of control over quality and relevance but is constrained by the diversity limitations and high generation costs~\citep{chang2024scaling,wang2022self}.

\begin{figure}[t]
    \centering
    \begin{subfigure}[b]{0.49\textwidth}
        \centering
        \includegraphics[width=\textwidth]{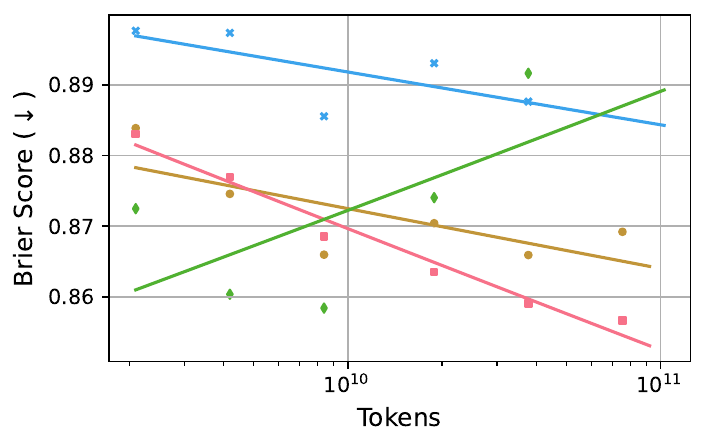} %
        \caption{Brier Score ($\downarrow$), Medical domain}
        \label{fig:sub1}
    \end{subfigure}
    \hfill
    \begin{subfigure}[b]{0.49\textwidth}
        \centering
        \includegraphics[width=\textwidth]{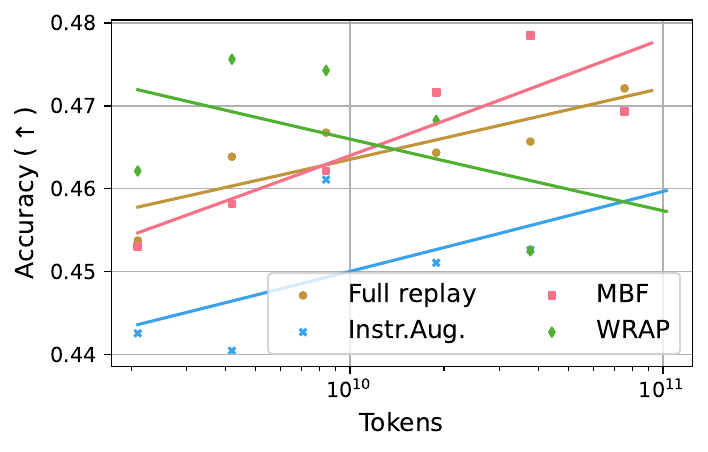} %
        \caption{Accuracy ($\uparrow$), Medical domain}
        \label{fig:sub2}
    \end{subfigure}
    \caption{
    Accuracy (right) and Brier Score (left) on MMLU Medical CF tasks for annealing experiments while upsampling domain-specific data. Each point represents the final performance of an independent run where 10\% of the training data was sampled from the corresponding method, and 90\% is the default training mix. The learning rate was linearly decayed to zero over the corresponding token budget.
    }
    \label{fig:2x2}
\end{figure}

Existing strategies for data sourcing and allocation decisions are often made ad hoc or based on single point estimates~\citep{olmo20242, grattafiori2024llama3herdmodels}. However, such estimates can be misleading. As shown in Fig.~\ref{fig:2x2}, in the low-compute regime, the synthetic data method WRAP~\citep{maini2024rephrasing} outperforms MBF in our experiments on the medical domain, but this relationship reverses as compute increases. This illustrates a key limitation of relying on point estimates when deciding which data source to scale: rankings between sources can shift dramatically with increased investment. The potential resource waste from such misguided decisions can be substantial — works like DeepSeek~\citep{liu2024deepseek,shao2024deepseekmath} have demonstrated the importance of synthetic data generation at scale, yet without proper scaling analysis there is a danger of substantial waste of resources. For instance, generating 100B tokens of synthetic data using a 70B parameter model could cost upwards of \$500K-\$1M in compute \footnote{Back-of-envelope: 70B params × 100B tokens × 2 FLOPs/param/token = $1.4\times1022^{22}$ 22 FLOPs. At \$2/A100-hour with 300 TFLOPs/sec throughput, this translates to roughly \$500K-\$1M in cloud compute costs.}, extensive model-based filtering can cost hundreds of thousands of dollars. Committing to a wrong strategy based on small-scale point estimates could thus waste hundreds of thousands of dollars in computational resources, highlighting the critical need for scaling-aware evaluation frameworks in data sourcing decisions. FLast but not least, we illustrate an example of a concrete practical use-case for out method in App.~\ref{app:prectical_scenario}. %

Another effective strategy for improving model performance is data source mixing ~\citep{ye2024data}. We would like to emphasize that data mixing is only possible when the data from the individual sources has already been collected, i.e. data mixing is performed posterior to committing to certain sources and spending the data mining budget. Hence, data mixing is not the focus of this work. Nevertheless, in App.~\ref{app:data_mix} we elaborate how the individual data source utility can potentially be used also for data mixing.

To address this, we propose to rely on domain-specific scaling laws instead of point-estimates in order to predict the utility of a data source. Overall, the contributions of this work are:
\begin{itemize}
    \item We demonstrate that data source rankings are not invariant across token scales, emphasizing the need for scaling-aware analysis when selecting data sources.

    \item We show that scaling curves can be constructed per data source, enabling better planning for data acquisition and compute allocation based on cost-utility trade-offs.

    \item To validate our approach we experiment with two domains—medical (well-represented in pretraining) and math (underrepresented)—using a 7B-parameter base model pre-trained for 1.2 trillion tokens. We evaluate multiple data acquisition strategies—including MBF, rephrasing techniques such as WRAP~\citep{maini2024rephrasing} or tiny-GSM ~\citep{liu2023tinygsm}, it's dialogue augmented version~\citep{olmo20242} and instruction augmentation~\citep{cheng2024instruction} with annealing runs ranging from 2B to 75B tokens. We show that our method enables data-driven decision making leading to more cost-effective model specialization.
\end{itemize}

\section{Methodology}
\subsection{Problem Setting}

\begin{figure}[t]
    \centering             
    \includegraphics[width=0.7\linewidth]{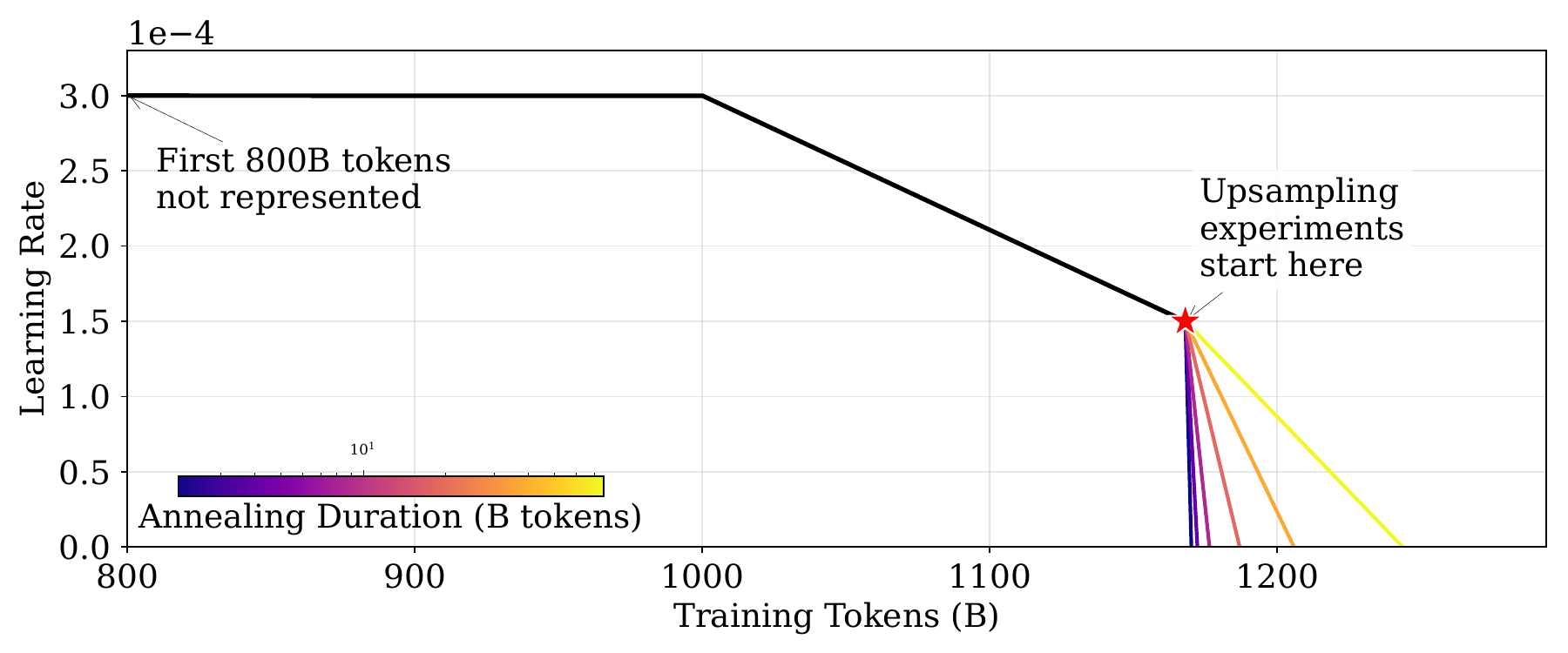}                   
    \caption{Learning rate schedule of our framework. Annealing w/ upsampling runs start from {\color{red}{$\star$}}.}
    \label{fig:lr_schedule}
\end{figure}  

We consider the problem of optimally allocating resources to data acquisition methods to maximize the downstream utility of the resulting dataset.

Data can be sourced from multiple distributions, denoted as \(\mathcal{D}_1, \mathcal{D}_2, \dots, \mathcal{D}_N\), each corresponding to a distinct acquisition method or a specific tuning of a method. Let \( c_i \) represent the cost of sampling a token from \( \mathcal{D}_i \), a collection of \( n \) such tokens form a dataset \( D_i(n) \).

Given a fixed budget \( C \) and a measure $\mathcal{U}(D)$ of the dataset's utility for a given task, our objective is to determine which data source maximizes $\mathcal{U}$, i.e.:

\begin{equation}
\label{eq:methodology_1}
    \underset{i}{\operatorname{argmax}}\quad \mathcal{U}(D_i(C \times c_i^{-1})),
\end{equation}
where $C \times c_i^{-1}$ denotes the maximum number of tokens that can be sampled from $\mathcal{D}_i$ within the budget $C$.

The central challenges lie in defining a function $\mathcal{U}$ that meaningfully captures the inherently underspecified notion of dataset utility and in estimating it at scale efficiently enough to inform data acquisition decisions.

In section \ref{subseq:data_utility}, we propose a specific formulation for \( \mathcal{U} \) and outline a scalable methodology for its estimation, facilitating informed data acquisition strategies.

\subsection{Dataset Utility Estimation via Annealing}
\label{subseq:data_utility}

We draw inspiration from Llama 3 \citep{grattafiori2024llama3herdmodels}, which evaluates domain-specific datasets by performing linear annealing over 40B tokens from a 50\%-trained 8B model. This approach efficiently extracts signal from challenging benchmarks using minimal compute, making it significantly more practical than training from scratch.

However, while the point estimates used in Llama 3 are effective for selecting fixed-size datasets, they do not capture how the utility of a data source scales with the number of tokens—a crucial consideration before committing substantial resources to data acquisition.

To address this limitation, we propose running short annealing experiments of varying durations, with a fixed 10\% upsampling of the data source under evaluation and 90\% of the default pretraining mix. After each run, we assess performance on a target task, favoring continuous metrics such as the Brier Score.

To isolate the contribution of the sourced tokens from the effect of extended training, we compute the difference in performance between each annealing run with upsampling and its equivalent run without upsampling:

\begin{equation}
\label{eq:methodology_2}
    \mathcal{U}(D)=S_{\text{base}}-S_{D}
\end{equation}
where $S_{D}$ denotes the performance metric (e.g., Brier Score or accuracy on a benchmark) after annealing the model while upsampling $D$ at a rate of $0.1$, and $S_{\text{base}}$ corresponds to the same annealing configuration without upsampling. If $D$ contains $n$ tokens, the annealing duration is set to $10n$ tokens, ensuring that each token in $D$ has been sampled once.

We compute $\mathcal{U}(D_i(n_i))$ for $n_i$ ranging from 210M to 7.5B tokens, and fit corresponding scaling laws. These scaling laws predict the utility of data source $\mathcal{D}_i$ at scale, enabling us to solve \eqref{eq:methodology_1} explicitly.

\section{Experiments}
We trained a baseline model with an architecture based on Mistral-7b~\citep{jiang2023mistral}. We start with a constant learning rate for 1T tokens followed by linear annealing over 336B tokens. We used a mixture of FineWeb-Edu~\citep{lozhkov2024fineweb-edu} and non-web part of the Dolma~\citep{dolma} dataset for pre-training. Our final checkpoint is close to the pareto frontier of the existing open souyrce models at the time of training: it reaches 56\% MMLU 5-shot accuracy, which is comparable to open-source models of similar size and training budget (as a reference, Zamba-7B \citep{glorioso2024zamba} reaches 57.7\% on the same benchmark with 1T pretraining tokens). We note, our goal here is not to train the best publicly available domain-specific model, but to propose a framework for estimating data source utility.
 All upsampling experiments initialize from the intermediate checkpoint at 168 billion tokens into annealing, where the learning rate had decayed to 50\% of its initial value (see Fig.~\ref{fig:lr_schedule}). Training details, including all hyperparameters and pretraining data, are provided in \ref{app:pretraining_details}. We selected the 7B scale for several reasons: (i) at the time of training, 7B models represented the upper bound of adopted practical scale for industrial applications, ensuring our findings would be relevant to real-world deployment scenarios; (ii) smaller models risk insufficient capacity to exhibit meaningful performance differences across data sources on challenging benchmarks~\citep{godey2024small}, potentially masking the scaling behaviors we aim to study; and (iii) while larger models might achieve domain adaptation through in-context learning alone, the computational cost of training multiple scaling runs at larger scales would be prohibitive for us.

\subsection{Data Acquisition Methods}

In the following, we describe the data acquisition methods used in this work. For each method, we aim to acquire a sufficient number of tokens such that no token repetition is necessary under the annealing hyperparameters outlined in Section~\ref{sec:annealing_exps}. %
Importantly, in our experiments, we match different data acquisition methods based on the number of unique upsampling tokens, rather than the compute cost of data curation. This choice is motivated by the observation that curation compute can vary significantly across methods. In a compute-matched setting, methods with higher curation costs—such as synthetic data generation—might produce too few tokens to yield meaningful signal on downstream tasks.

\texttt{Full replay} --- the annealing run is performed on the same data as the initial pre-training.

\texttt{MBF} --- model-based filtering (MBF) uses a BERT-regressor as quality filter, that was trained on 500k examples annotated by Meta-Llama-3-70B-Instruct \citep{llama3modelcard}. Several recent works showed that using such trained quality classifier can lead to substantial improvements of the downstream performance \citep{fang2023data,lozhkov2024fineweb-edu,soldaini2024dolma,li2024datacomp,olmo20242}. We present additional details and prompts used for training set annotation in App.~\ref{app:mbf}.

\texttt{WRAP} --- Web Rephrase Augmented Pre-training (WRAP) proposed by \citet{maini2024rephrasing} relies on rephrasing the pre-training data using different language and style (e.g. ''like Wikipedia``).\citet{maini2024rephrasing} shows that such rephrasing can lead to faster learning in the pre-training phase. We follow the original work and include rephrasing in three styles: scholar language, Wikipedia style and Q/A. We additionally add a rephrasing in Q/A style that is close to MMLU format which led to significant improvements on multiple-choice tasks in the MMLU multiple-choice (MC) format. %
Due to high cost of WRAP, our longest annealing run for this method only contained 3.8B tokens (18,000 annealing steps). We elaborate further details of this method in App.~\ref{app:wrap}.  %

\texttt{Instr. Aug.} --- we experiment with augmenting a subset of highly scored \texttt{MBF} documents with instruction format as proposed by \citet{cheng2024instruction}. We use the pre-trained 8B instruction synthesizer and code released by \citet{cheng2024instruction} to augment selected seed documents with generated tasks. Augmenting pre-training data with downstream tasks data or NLP tasks has been shown effective in a number of recent works \citep{cheng2023adapting,krishna2022downstream}.

For the math domain we consider synthetic data generating methods specialized on the math domain.

\texttt{TinyGSM} --- \citet{liu2023tinygsm} proposed to augment the training set of the original GSM8k \citep{cobbe2021gsm8k} dataset with synthetically generated problems and python solutions using GPT3.5 model. This augmentation resulted in a synthetic dataset containing 1.8B tokens. In App.\ref{app:math_cost} we elaborate how we estimate the curation cost for this dataset.

\texttt{TinyGSM-MIND} ---- \citet{olmo20242} further improved the quality and diversity of the TinyGSM by filtering out samples with non-executable code and rephrasing the remaining problems in the style optimized for the math domain --- MIND style \citet{akter2024mind}, using Qwen2.5-7B-Instruct model.

In order to study the importance of the formatting (see Section \ref{sec:domain_eval_metrics} and Fig.~\ref{fig:format_importance_mc}), we introduce the following two baselines:
\texttt{WRAP+Q/A (Wiki)} uses Wikipedia articles extracted from Dolmino \citep{olmo20242}, unrelated to the medical domain, and augments them with MMLU-style Q/A. \texttt{WRAP (w/o mmlu-Q/A)} is the same as \texttt{WRAP} but without the MMLU-style Q/A.

We provide the details of compute estimation for various methods in App.~\ref{sec:cost_estimation}, where we adopt the $2 \times |P|$~\citep{kaplan2020scaling} approximation\footnote{More precise estimates can be found in \citet{scaling-book}} of inference FLOPs per token, with $|P|$ denoting the number of parameters of the inference model.

\subsection{Domain and Evaluation Metrics}
\label{sec:domain_eval_metrics}
We focus our experiments on two target domains: 

\textbf{Maths}, where high-quality data is relatively scarce in the pretraining corpus -- only ~10\% of the FineWeb-Edu dataset received a score above 2.5 from the Math MBF classifier -- resulting in poor performance of the base model: $\sim$~33 \% on MMLU-maths.

\textbf{Medical}, where high-quality data is more abundant -- ~28\% of FineWeb-Edu samples scored above 2.5 by the Medical MBF classifier -- leading to better performances: $\sim$~56\% on MMLU-medical.%

We adopt the Brier Score \citep{brier1950verification} ($\downarrow$) as our primary metric for multiple-choice tasks. This choice is motivated by \citet{schaeffer2023emergent}, who argues that switching from a discontinuous metric like accuracy to a continuous one like Brier Score can more effectively reveal emergent behaviors in LLMs, making it more suitable for scaling law estimation. For some math and medical tasks, we use Exact Match ($\uparrow$) as detailed in Table~\ref{tab:eval_tasks}.

We report metric deltas, \texttt{Metric} $\Delta$, such as \texttt{Brier Score $\Delta$}, which represent the difference between the metric's value for the full replay baseline and the given model's metric value, corresponding to the utility function in Equation~\ref{eq:methodology_2}. Thus, a \texttt{Brier Score $\Delta$} below zero indicates better performance than the full replay baseline, while for \texttt{Exact Match $\Delta$}, higher values indicate superior performance. Most of the plots presented here use a log-log scale to better reflect the power-law nature of the scaling laws. Intuitively, the \texttt{Metric} $\Delta$ gives and indication of how much \texttt{Metric} has changed as a consequence of upsampling data, indicating the net benefit of the data acquisition.

\begin{figure}[t]
    \centering
    \begin{minipage}[t]{0.6\linewidth}
        \centering
        \includegraphics[width=\linewidth]{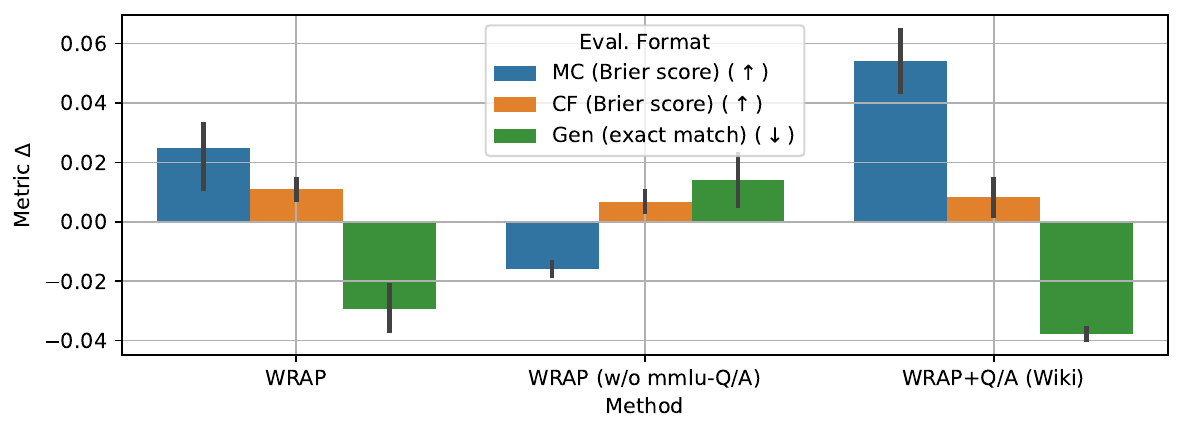}
        \caption{Impact of WRAP version across evaluation formats. The y-axis shows the difference to full replay baseline (0 means same as full replay) for MC ($\uparrow$), CF ($\uparrow$) and generative formats ($\downarrow$) on medical MMLU tasks. Performance is averaged over runs of 1, 2 and 4 thousand steps. %
        }
        \label{fig:format_importance_mc}
    \end{minipage}%
    \hfill
    \begin{minipage}[t]{0.38\linewidth}
        \centering
        \includegraphics[width=\linewidth]{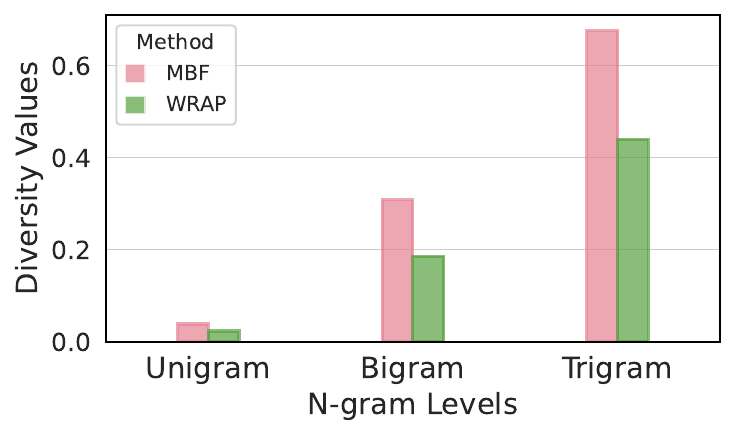}
        \caption{Comparison of \texttt{MBF} and \texttt{WRAP} methods in terms of Distinct N-gram Scores. \texttt{MBF} consistently shows higher diversity than WRAP across all N-gram levels. %
        }
        \label{fig:n-gram-diversity}
    \end{minipage}
\end{figure}

For evaluation, we rely on the LM Evaluation Harness library \citep{eval-harness}. We select tasks related to the medical and math domains in both Multiple Choice (MC) format, CF\footnote{CF format is named continuation in LM Evaluation Harness.} and the generative version of the task's. We primarily adopt the CF style in our experiments. This choice is motivated by our observation that CF is significantly less sensitive to the format of pre-training data compared to MC and generative formats. We illustrate this point in Fig.~\ref{fig:format_importance_mc}, which shows that removing MMLU-style Q/A from \texttt{Wrap} --- \texttt{Wrap (w/o mmlu-Q/A}), results in a large performance drop in MC and generative tasks, in both cases degrading performances~\footnote{\citet{muennighoff2024olmoeopenmixtureofexpertslanguage} finds CF to provide much stronger signal during pre-training than the MC version of the task.}. In comparison, the performance on CF formatted tasks remains consistent (and better than baseline) across all three formatting versions. Additionally, taking unrelated Wikipedia documents and augmenting them with MMLU-style Q/A -- \texttt{Wrap+Q/A (Wiki)} -- results in a large improvement in MC and generative evaluations, without visible effect on CF. This suggests that CF is a more robust and format-invariant evaluation strategy. The full list of tasks used, organized by domain and evaluation format, is provided in Table~\ref{tab:eval_tasks}.

\subsection{Annealing Experiments}
\label{sec:annealing_exps}
We perform annealing runs at 1, 2, 4, 9, 18, and 36 thousand annealing steps (from 2.1B up to 75B tokens). Each run uses a linear learning rate schedule, starting from the first stage's learning rate ($1.515 \times 10^{-4}$), with the learning rate linearly decayed over the number of annealing steps for each run. We use a batch size of 256 and a sequence length of 8192 tokens per sample. Evaluations are conducted at the end of each annealing run.

All experiments are conducted with a replay ratio of 90\%, meaning that approximately 10\% of the examples in each mini-batch come from the upsampled target domain. This ratio was selected based on a hyperparameter search conducted on MBF data in the medical domain and was held constant for the remainder of the experiments. All annealing runs are conducted using the Fast-LLM framework~\citep{Lamy_Poirier_Fast_LLM_2024}\footnote{We mostly use the \texttt{sha-ff1486d} version for the annealing runs}. We run two seeds and average the results for the full replay baseline, and only use a single seed for other baselines to minimize compute cost.

\section{Results and Analysis}

\begin{figure*}[t]
    \centering
    \begin{subfigure}{0.49\linewidth}
        \centering
        \includegraphics[width=\linewidth]{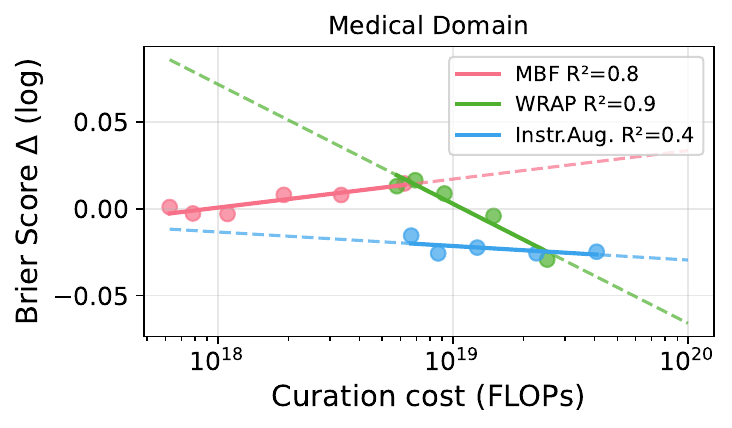}
        \label{fig:brier_score_delta_med_curation_only}
    \end{subfigure}
    \hfill
    \begin{subfigure}{0.49\linewidth}
        \centering
        \includegraphics[width=\linewidth]{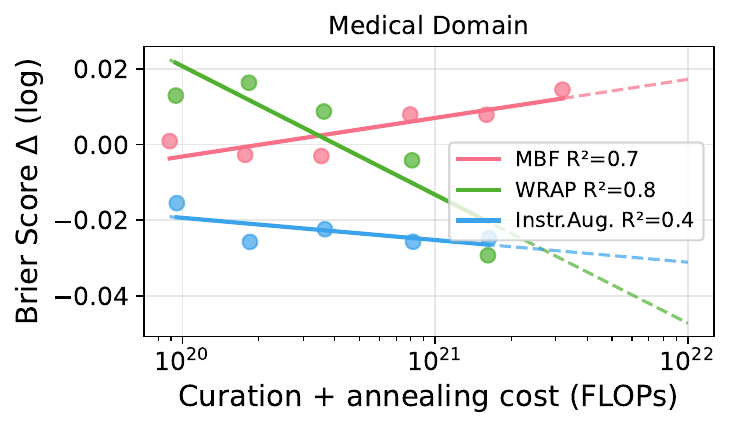}
        \label{fig:brier_score_delta_med_full_compute}
    \end{subfigure}
    \centering
    \begin{subfigure}{0.49\linewidth}
        \centering
        \includegraphics[width=\linewidth]{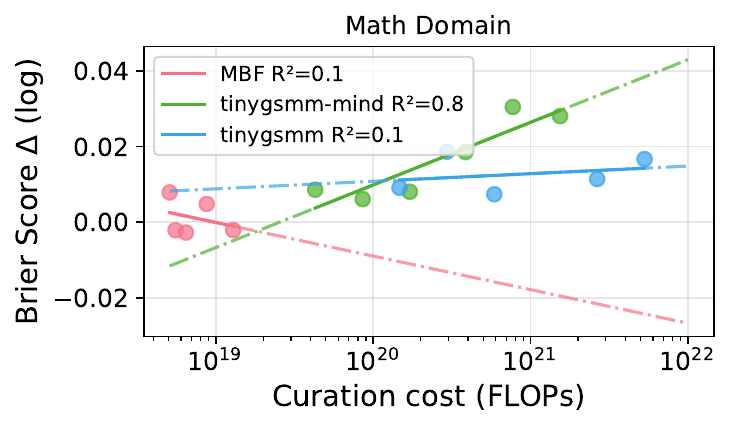}
        \label{fig:brier_score_delta_math_curation_only}
    \end{subfigure}
    \hfill
    \begin{subfigure}{0.49\linewidth}
        \centering
        \includegraphics[width=\linewidth]{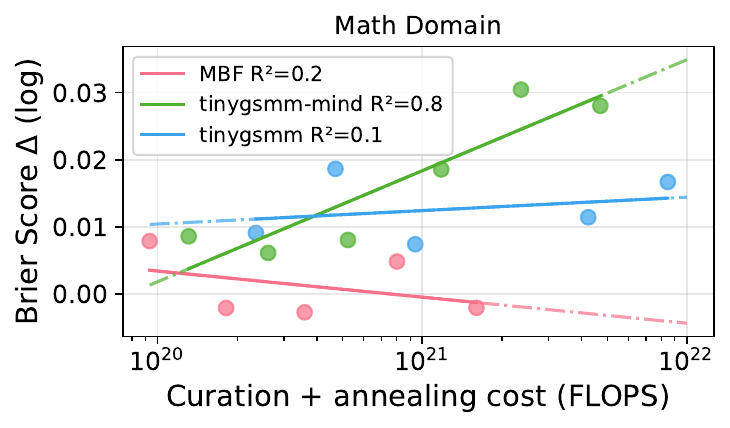}
        \label{fig:math_brier_delta_total_cost}
    \end{subfigure}
    \caption{Brier Score $\Delta$ to full replay ($\uparrow$) on medical (top) and math (bottom) MMLU tasks in the CF format vs. curation only (left) and curation + annealing (right) compute cost (FLOPs). The dotted part of the lines are extrapolated.}
    \label{fig:math_and_medical_brier}
\end{figure*}

\subsection{Scaling Trends and Cost Efficiency}
In Fig.~\ref{fig:math_and_medical_brier}, we analyze the scaling behavior of different data sourcing methods in the medical and math domains. We present two alternatives of the cost function $c_i$: one that only considers the cost of sampling from distribution $\mathcal{D}_i$, and one that also accounts for the cost of the annealing training steps, effectively adding a constant cost per token to all methods. Focusing solely on curation costs implies treating dataset acquisition as a distinct budget, separate from pretraining—an approach well-suited for datasets intended for reuse across multiple models. Conversely, jointly optimizing curation and annealing compute costs accounts for scenarios where a single budget must be allocated between data acquisition and adaptation steps, aiming to maximize final model performance within a fixed computational constraint.

At smaller compute scales, the synthetic \texttt{WRAP} method outperforms the quality-filtered \texttt{MBF} data. However, as compute investment increases, we observe diminishing returns from \texttt{WRAP} and steadily increasing utility from \texttt{MBF}. A similar observation has been made by \citet{chang2024scaling} at a much smaller model scale, where they observed that synthetic data has higher utility at smaller compute. This highlights a key limitation of relying on point estimates from low-compute regimes, which would incorrectly favor \texttt{WRAP} over \texttt{MBF}. In contrast, our approach—grounded in scaling law estimation—reveals the long-term advantages of \texttt{MBF}, enabling more informed data source selection. We observe a similar, yet less pronounced effect on the math domain (bottom of Fig.~\ref{fig:math_and_medical_brier}), where \texttt{TinyGSM} tends to performs better than \texttt{MIND} at small compute budgets, yet \texttt{MIND} scales better overall. Results on the math domain also highlight that synthetic data can be made diverse and scale effectively, which partially contrasts the observations of~\cite{chang2024scaling}.

We hypothesize that the bad scaling of \texttt{WRAP} on the medical domain is due to low diversity, as suggested by Fig.~\ref{fig:n-gram-diversity} and discussed in further depth in Fig.~\ref{fig:entropy}. While the high quality of \texttt{WRAP} gives it the advantage at small scales, this redundancy eventually make the upsampling of its tokens hurtful after a certain scale, which is effectively predictable from our observations below that threshold. While initially less impressive, \texttt{MBF} reliably improves the utility of the data as the sampling size increases.

Surprisingly, we find that instruction augmentation does not outperform full replay on CF tasks (Fig.~\ref{fig:math_and_medical_brier}). However, it proves as effective as WRAP on MC formatted downstream evaluations (Fig.~\ref{fig:med_scalling_curves_mc}). This suggests that instruction augmentation primarily enhances benchmark performance through formatting rather than improving the model’s underlying knowledge, and aligns with the findings of Fig.~\ref{fig:format_importance_mc}: the MC format is suboptimal for assessing knowledge in LLMs due to its sensitivity to formatting data.

\subsection{Effectiveness of Different Data Sources}
\begin{wrapfigure}[15]{r}{0.6\textwidth}
    \centering             
    \includegraphics[width=0.8\linewidth]{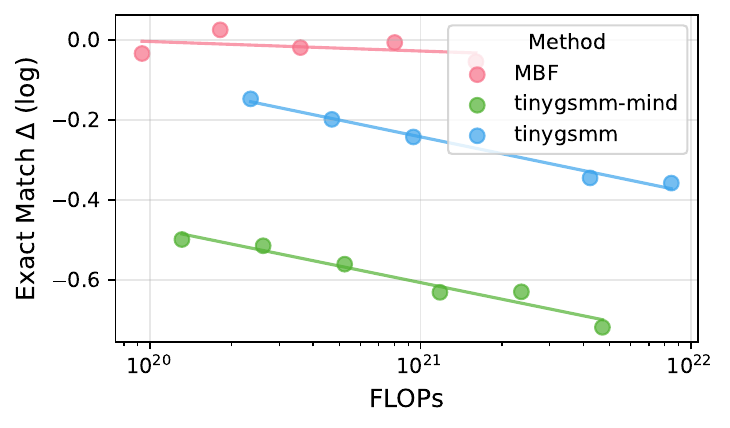}                   
    \caption{Exact match $\Delta$ (log) to full replay ($\downarrow$) on non-MMLU math tasks vs. compute (FLOPS).}
    \label{fig:exact_maych_delta_math}
\end{wrapfigure}
While different data sources are often tailored to specific domains, our experiments reveal that their effectiveness also varies significantly depending on the downstream evaluation format and metric. For instance, as previously discussed, instruction-based augmentation proves more effective on tasks evaluated in MC format. Similarly, we observe that the \texttt{WRAP} method performs better on MC tasks, as shown in Fig.~\ref{fig:med_scalling_curves_mc}. This can be attributed to the fact that our version of \texttt{WRAP} augments the data using MMLU-style question/answer pairs, as further supported by Fig.~\ref{fig:format_importance_mc} and discussion in Section~\ref{sec:domain_eval_metrics}. In the math domain, we see a parallel pattern: both \texttt{TinyGSM} and its \texttt{MIND} variant yield measurable improvements primarily on tasks evaluated in the CF format, as demonstrated in Fig.~\ref{fig:math_scalling_curves_cf}. Furthermore, Fig.~\ref{fig:exact_maych_delta_math} shows that these two data sources lead to substantial performance gains on non-MMLU math tasks, particularly when evaluated using exact match metrics.

These results highlight the importance of jointly considering the evaluation format and the nature of the data source when designing or selecting metrics and datasets for data source evaluation. These seemingly innocuous variations can cause significant variations in the relative performance of data collection methods across similar-looking tasks. This low transferability of method performance further motivates using a quantitative framework to guide task-specific data acquisition decisions.

\subsection{Limitations} 

At smaller compute budgets, our performance estimates can be influenced by the stochasticity of batch sampling. While most data sources exhibit robust scaling trends, certain methods, such as WRAP, show greater variability at low scales, leading to outliers that can distort scaling law coefficients. This issue could be mitigated by averaging results over multiple random seeds, particularly at lower scales, though at the cost of additional compute. However, we find that despite this variability, the overall trends remain consistent at the scales we study, allowing us to reliably infer method rankings at the highest scale from lower-scale experiments.

Furthermore, due to computational constraints, we were unable to conduct extensive ablations. For example, it would be valuable to analyze how utility scaling behaves for a single data source across a range of tunings, or to assess the sensitivity of our framework to the upsampling ratio and to the choice of initial checkpoint—particularly the impact of its starting learning rate. Additionally, we could not extend our annealing and upsampling experiments to scales orders of magnitude larger than ours, leaving open the question of how well our derived scaling laws generalize across vastly larger compute budgets.

\section{Related Work}
\textbf{Domain-Specific Data Acquisition:} Domain-specific data acquisition has emerged as a more effective strategy for pre-training language models in specialized fields, as targeted collections of relevant content consistently yield better performance than massive but unfocused Internet datasets \citep{hwang2025subset, parmar2024data, dong2024once}. Some recent work shows the effectiveness of targeted data collection and synthetic data generation as two effective ways to improve model performance. \citet{shao2024deepseekmath} employs an iterative step-by-step approach that combines automated filtering with human validation. In order to increase mathematical reasoning capabilities of the model, the authors curated a 120B token dataset rich in mathematical content, which also involved training a FastText classifier on a seed dataset to identify "math-like" content within Common Crawl, followed by human annotation to ensure data quality and relevance. Although targeted data acquisition has downstream utility, it can be computationally expensive. \cite{bansal2024smaller} shows that, under fixed compute budgets, sampling data from weaker but cheaper models can yield more diverse and effective training data than relying solely on stronger, more expensive models. \citet{adler2024nemotron} releases an open-source synthetic data generation pipeline as part of the release of Nemotron-340B parameter models. These models facilitate the creation of high-quality domain-specific training data, addressing challenges related to data scarcity. While these approaches demonstrate the potential of various data acquisition strategies, there is a lack of methods for comparing their effectiveness at different scales. Our work addresses this gap by proposing a scaling law framework that enables practitioners to quantitatively evaluate and compare the utility of different data sources.

\textbf{Dataset Utility Estimation:} Recent works have explored various approaches to optimize data mixtures (data mixtures can be seen as a seperate source in our framework) for LLM pre-training. Notably, RegMix \citep{liu2024regmix} proposes formulating data mixture selection as a regression task, training many small models (1M parameters) on diverse mixtures to predict the performance of unseen combinations, then applying the best mixture to train larger models (1B parameters). %
Similarly, \cite{olmo20242} employs a mid-training curriculum approach called "micro-annealing", where small batches of quality-assessed data validate the effectiveness of the model in specific datasets. Other works have focused on data ablation approximations through parameter averaging of models trained on different partitions, allowing efficient evaluation of various data mixtures without expensive joint training \citep{na2024scalable}. In contrast to these point-estimate approaches, scaling law methods provide a more comprehensive framework. \cite{goyal2024scaling} demonstrates that data curation cannot be compute-agnostic, as high-quality filtered data rapidly loses utility when repeated, eventually requiring inclusion of "unseen" but "lower-quality" data. These scaling laws characterize the differing utility of various data subsets and explain how utility diminishes with repetition. ScalingFilter \citep{li2024scalingfilter} leverages the perplexity difference between models of different sizes as a quality indicator, inversely utilizing scaling laws to curate high-quality datasets without relying on reference data. Our work extends these approaches by estimating scaling laws at a larger scale, focusing on the utility estimation of the dataset rather than the annealing phases or domain specialization.

\textbf{Data Allocation Strategies:}
Advances in data allocation strategies have demonstrated the effectiveness of dynamic, scaling-law-driven approaches for optimizing data mixtures. Adaptive Data Optimization (ADO) \citep{jiang2024adaptive} eliminates reliance on proxy models by leveraging per-domain scaling laws to dynamically adjust data distributions during training, enabling computationally efficient optimization without interrupting model updates. Complementing this, \cite{ye2024data} introduce Data Mixing Laws, which seeks quantitative predictability of model performance across mixtures through functional relationships, allowing scaling law extrapolations to predict optimal proportions for large-scale training with minimal experiments. These methods advance beyond static heuristics or point estimations used in earlier approaches like DoReMi \citep{xie2023doremi}. Moreover, \cite{agarwal2025deliftdataefficientlanguage} introduces DELIFT, an approach to do data-efficient fine-tuning of large language models by employing a versatile pairwise utility metric combined with submodular optimization techniques for optimal data selection. These approaches demonstrate that going beyond point estimates in mixture optimization can enable more efficient data allocation strategies, crucial for both pretraining and fine-tuning regimes in LLMs.  In contrast to these data mixing approaches, our work focuses on the preceding question of evaluating individual data sources before mixture optimization — providing the utility estimates that inform which sources are worth including in downstream mixing strategies.

\section{Conclusion}
In this work, we introduced a practical method for estimating the value of different data sources when adapting a pre-trained language model to specific domains. Rather than relying on single-point evaluations or small-scale training runs, we leveraged multiple short annealing runs to construct scaling curves that predict performance variations as a function of compute. This approach mitigates the risk of misleading conclusions, particularly in cases where the relative ranking of data sources shifts with scale.

We applied our method to two domains: medical and math. Our experiments showed that some data sources, like model-based filtering, can become more effective as compute increases, while others, like synthetic data (e.g., \texttt{WRAP}) can be sometimes more useful at smaller compute budgets but suffer from severe diminishing returns.

By comparing both the training and data generation costs, we showed the importance of these trade-offs when making data acquisition decisions. Our results highlight the importance of matching data sources not only to the domain, but also to the evaluation format and available compute. Overall, our methodology can lead to more informed and cost-effective strategies for domain-specific pretraining.

Finally, because any mixture of data sources can itself be treated as a data source, our approach naturally extends to optimizing data mixtures by evaluating different candidate combinations. Unlike the standard practice of deriving scaling laws over model size to guide mixture selection~\citep{ye2024data,grattafiori2024llama3herdmodels}, our method enables predictions from a relatively small number of sampled tokens. This not only reduces computational cost but also reveals meaningful signal on benchmarks where smaller models might otherwise be saturated.

\newpage

\bibliography{bib_colm2025_conference}
\bibliographystyle{colm2025_conference}

\newpage
\appendix
\onecolumn
\section{Practical Scenario}
\label{app:prectical_scenario}
Note, the practical scenario we aim to address here is when practitioners face the important decision about which specific data source they should invest into. Practical example: a pharmaceutical company wants to improve their LLM for drug discovery. They can evaluate progress on public benchmarks, but need billions of tokens for effective annealing - far more than their proprietary datasets contain. Should they invest in filtering PubMed papers, generating synthetic chemical data with GPT-4, or purchasing expensive databases? Our framework helps decide which data acquisition strategies are worth pursuing before committing too many resources to any given source. We also highlight, that this decision has to be made before data mixing is possible, i.e. for data mixing the data of different sources must already be available.

\section{Data Mixing}
\label{app:data_mix}
Here we sketch how data source specific scaling laws can be used to estimate optimal data mixing coefficients. 

Given fitted utility scaling laws per data source $i$ of the form $\Delta_i(c_i) = a_i + b_i \log(c_i)$, where $\Delta_i$ is the utility improvement (e.g., reduction in Brier score), and $c_i$ is the compute budget allocated to source $i$, the total gain from a mixture can be approximated (assuming additive independence of source utility) as:
$\Delta_{\text{mixture}} = \sum_{i=1}^n (a_i + b_i \log(c_i))$, s.t. $\sum_{i=1}^n c_i = c_{\max}$, $c_i \ge 0$, where $c_max$ is the maximum commute budget. This is a constrained concave maximization problem (or minimization in case of Brier score) and has a closed form optimality conditions, i.e. $c_i = \frac{b_i}{\sum_j b_j} \, c_{\max}$. While we leave this direction for future work, this can potentially yield a simple easily implementable rule for data source mixing where the data source weights in the mix are allocated proportionally to the slopes of the individual scaling lows $b_i$. 

\section{Training details}
\subsection{Base Model and Pretraining Data}
\label{app:pretraining_details}

\textbf{Training procedure:} Our baseline model is based on the architecture of Mistral-7b~\citep{jiang2023mistral} and uses the same tokenizer. It is trained with AdamW~\citep{loshchilov2017decoupled}, using a sequence length of 8192 tokens and 256 sequences par minibatch, for a total of 2.1M tokens. We use $\beta_1=0.9$ and $\beta_2=0.95$ as first and second moments, respectively.
The training is done in mixed precision over three stages: we first warmup the model by increasing linearly the learning rate to $3e^{-4}$ over 2000 steps. Then, we use a constant learning rate of $3e^{-4}$ for 478k steps. Finally, we anneal linearly to zero learning rate over 160k steps. The base checkpoint for the experiments presented in this work corresponds to the 80,000th step of the annealing, when the learning rate has reached $1.5e{-4}$. This corresponds to a total of 560k iterations, i.e., 1.18T training tokens. We use FastLLM~\citep{Lamy_Poirier_Fast_LLM_2024} as training engine with FlashAttention 2~\citep{dao2023flashattention2} and ZeRO stage 3~\citep{rajbhandari2020zeromemoryoptimizationstraining}, and train the model on 64 H100 GPUs with full data parallelization, for a total duration of 32,500 H100-hours, averaging 10,000 tokens/s/GPUs.

\textbf{Default Pretraining Mix:} Our pretraining dataset is the concatenation of the Dolma~\citep{dolma} dataset from which the Common Crawl subset has been removed, and Fineweb-edu~\citep{lozhkov2024fineweb-edu}.

\section{Data acquisition methods}
\subsection{Mode-based filtering details}
\label{app:mbf}

We followed a strategy similar to ~\cite{lozhkov2024fineweb-edu} for training our math and medical classifiers. We began by designing prompts to annotate high-quality documents in each domain and used these annotations to train classifiers for filtering. After iterating on several prompt variations we landed on the prompts following prompts for math and medical respectively~\ref{fig:math_prompt} and \ref{fig:medical_prompt}.~\footnote{We iterated on couple of prompts inspired by ~\citep{lozhkov2024fineweb-edu} then annotated 100K fineweb-edu documents using Llama3-70B we used these annotations to build a classifier, filter, and performed (up-sampled) annealing experiments. Based on the performance we picked the best prompt among the candidates.} 
For the final classifier, we used 500K annotations from Llama3-70B. We also conducted ablations on classifier training, comparing binary classification with regression and exploring up-sampling vs. down-sampling in the medical domain. Regression performed best in annealing experiments, leading us to adopt it for the math domain as well.
Our experiments revealed that MBF is sensitive to the classifier threshold. In our ablations, we tested using only the top-K highest-scoring documents but found that annealing them performed worse than replay. This suggests that a lack of diversity among top-K documents degrades model performance.~\footnote{We computed perplexity scores under the base model and found that top-K documents had lower scores.} To address this, we conducted a parameter sweep for the classifier threshold, ranging from 2 to 5 in 0.5 increments, and found that a threshold of 2.5 yielded the best performance on downstream tasks. Hence, unless stated otherwise, for \texttt{MBF} we apply the filtering threshold of 2.5 which we ablated on the medical domain. This is similar to what has been been used by \citep{olmo20242}, who used the threshold of 3.

\begin{figure*}[b]
    \begin{framed}
    \ttfamily %
        Evaluate an extract for its value in presenting mathematical information, use the following additive 5-point scoring system. Points are awarded based on the satisfaction of each criterion: 
        \begin{itemize}
            \item Award a point if the extract contains some mathematical information, terminology, or references to mathematical concepts, even if it includes irrelevant content such as advertisements, promotional material, job posts, or non-academic details. The mathematical information should still be accurate and relevant.
            \item Add a second point if the extract touches on general mathematical topics or some calculations but is disorganized, unclear, or lacks depth. It may include a mix of relevant and irrelevant information, making it less effective for structured understanding.
            \item Award a third point if the extract provides coherent and accurate mathematical information suitable for general use. It may offer clear explanations of theories, formulas, or mathematical principles, though it could include some advanced terms or concepts that require further clarification. The extract should be appropriate for students, educators, or general audiences.  
            \item Grant a fourth point if the extract is highly relevant and well-organized, presenting clear and detailed mathematical information such as problem-solving strategies, theoretical insights, or applied mathematics examples. The content should be coherent, with minimal unrelated material, and it should be useful for mathematicians, educators, or individuals seeking in-depth mathematical knowledge. Complex terminology may be used, but it should be contextually explained. 
            \item Bestow a fifth point if the extract is outstanding in its clarity, depth, and relevance to mathematical topics. It should present comprehensive and well-researched information with detailed insights into mathematical theories, advanced concepts, or applied mathematics. The content should be precise, devoid of unnecessary details, and offer profound value to mathematicians, researchers, or those seeking expert-level information.
        \end{itemize}
        \normalfont %
    \end{framed}
     \caption{5-Point scoring prompt for math}
     \label{fig:math_prompt}
\end{figure*}

\begin{figure*}[b]
    \begin{framed}
    \ttfamily %
        Evaluate the following extract for its value in presenting medical or health-related information. Use the additive 5-point scoring system described below. Points are awarded based on the satisfaction of each criterion:
        \begin{itemize}
            \item Add 1 point if the extract provides some medical/health information or includes any medical/health related jargons, even if it includes irrelevant content such as advertisements promotional material, job posts or non-academic details. The medical or health information should still be accurate and relevant.
            \item Add another point if the extract touches on general biology, health or medical topics, but the presentation is disorganized, unclear, or lacking in detail. It may include a mix of relevant and non-relevant information, making it less effective for structured understanding.
            \item Award a third point if the extract provides coherent and accurate medical or health-related information that is suitable for general use. It may offer clear explanations of treatments, diagnoses, or research findings, though it could include some advanced terms or concepts that require further clarification. The extract should be appropriate for health professionals, students, or general audiences.  
            \item Grant a fourth point if the extract is highly relevant and well-organized, presenting clear and detailed medical information such as treatment protocols, research summaries, or clinical guidelines. The content should be coherent, with minimal unrelated material, and it should be useful for practitioners or individuals seeking in-depth medical knowledge. Complex medical terminology may be used, but it should be contextually explained. 
            \item Bestow a fifth point if the extract is outstanding in its clarity, depth, and relevance to medical or health-related topics. It should present comprehensive and well-researched information with detailed insights into treatments, clinical practices, or recent research findings. The content should be precise, devoid of unnecessary details, and offer profound value to healthcare professionals, researchers, or those seeking expert-level information.
        \end{itemize}
    \normalfont %
    \end{framed}
    \caption{5-Point scoring prompt for medical}
    \label{fig:medical_prompt}
\end{figure*}

\subsection{WRAP}
\label{app:wrap}
Unless stated otherwise, we use a randomly selected subset of 1 million (2.25 billion tokens) highly-scored MBF document ($\geq$ 5) as seed texts for \texttt{WRAP} and use Meta-Llama-3.2-3B-Instruct~\citep{llama3modelcard} as out synthesis model. For the medical domain this resulted in generation of around 2.78 billion new tokens resulting in the total of around 5 billion \texttt{WRAP} tokens. We note that this is significantly lower than the number of tokens needed for our longest annealing run, which requires 7.5 billion unique up-sampled tokens. Following \citep{maini2024rephrasing} we use include rephrasing in three styles: scholar language (Fig.~\ref{fig:wrap_scholar}), Wikipedia (Fig.~\ref{fig:wrap_wiki}) style and Q/A (Fig.~\ref{fig:wrap_qanda}). We additionally add a rephrasing in MMLU-like Q/A style (Fig.~\ref{fig:wrap_mmluqa}).

\begin{figure}[b]
    \begin{framed}
    \ttfamily %
            For the following document give me a diverse paraphrase of the same in high quality English language as in sentences on Wikipedia. 
            Output the paraphrase directly, do not include any other text.
            Document:

            \{document\}
    \end{framed}
     \caption{WRAP Scholar style prompt.}
     \label{fig:wrap_scholar}
\end{figure}

\begin{figure}[b]
    \begin{framed}
    \ttfamily %
            Convert the following document into a conversational format with multiple tags of "Question:" followed by "Answer:". 
            Output the conversation directly, do not include any other text.
            Document: 

            \{document\}
    \end{framed}
     \caption{WRAP Q\&A style prompt.}
     \label{fig:wrap_qanda}
\end{figure}

\begin{figure}[b]
    \begin{framed}
    \ttfamily %
            Here are \{qa\_n\_shot\} question-answer pairs based on a document:
            \{context\_qa\_pairs\}

            Below is a new document. Based on the style and format of the previous question-answer pairs, generate as many high-quality question-answer pairs as you can about the content of the document.
            Output the new question-answer pairs directly, do not include any other text.
            Document:            
            \{document\}
    \end{framed}
     \caption{WRAP MMLU-style Q\&A prompt. Here \texttt{context\_qa\_pairs} are the in-context examples randomy sampled from MMLU validation set.}
     \label{fig:wrap_mmluqa}
\end{figure}

\begin{figure}[b]
    \begin{framed}
    \ttfamily %
            For the following document give me a paraphrase of the same using very terse and abstruse language that only an erudite scholar will understand.
            Replace simple words and phrases with rare and complex ones. 
            Output the paraphrase directly, do not include any other text.
            Document:
            
            \{document\}
    \end{framed}
     \caption{WRAP Wikipedia style prompt.}
     \label{fig:wrap_wiki}
\end{figure}

\begin{table*}[t]
    \centering
    \small
    \begin{tabularx}{\linewidth}{l|>{\hsize=1.5\hsize}X>{\hsize=0.5\hsize}X}
        \toprule
        \textbf{} & \textbf{Tasks} & \textbf{Metric} \\ 
        \midrule
        Medical MMLU CF tasks & ``mmlu\_anatomy``, ``mmlu\_clinical\_knowledge``, ``mmlu\_college\_biology``, ``mmlu\_college\_medicine``, ``mmlu\_high\_school\_biology``, ``mmlu\_medical\_genetics``, ``mmlu\_professional\_medicine`` & Brier Score ($\downarrow$) \\ 
        \hline
        Medical MMLU MC tasks & Same as Medical MMLU CF tasks but in MC format & Brier Score ($\downarrow$)\\
        \hline
        Medical MMLU Generative tasks & Same as Medical MMLU CF tasks but in generative format & Exact match ($\uparrow$)\\
        \hline
        Medical MC tasks & Medical MMLU MC tasks + ``pubmedqa``, ``medqa\_4options``, ``medmcqa`` & Brier Score ($\downarrow$) \\
        \midrule
        \midrule        
        Math MMLU-tasks CF & 
       ``mmlu\_continuation\_abstract\_algebra``,
       ``mmlu\_continuation\_college\_mathematics``,
       ``mmlu\_continuation\_elementary\_mathematics``,
       ``mmlu\_continuation\_high\_school\_mathematics``,
       ``mmlu\_continuation\_high\_school\_statistics`` & Brier Score ($\downarrow$) \\
        \hline
        Math non-MMLU tasks & ``gsm8k\_cot``, ``hendrycks\_math\_algebra``,
       ``hendrycks\_math\_counting\_and\_prob``, ``hendrycks\_math\_geometry``,
       ``hendrycks\_math\_intermediate\_algebra``, ``hendrycks\_math\_num\_theory``,
       ``hendrycks\_math\_prealgebra``, ``hendrycks\_math\_precalc``  & Exact match ($\uparrow$) \\
        \bottomrule
    \end{tabularx}
    \caption{Tasks used for evaluation in the medical and math domains. CF refers to continuation format, and MC to multiple choice format.}
    \label{tab:eval_tasks}
\end{table*}

\begin{figure*}
    \centering
    \begin{subfigure}{0.49\linewidth}
        \centering
        \includegraphics[width=\linewidth]{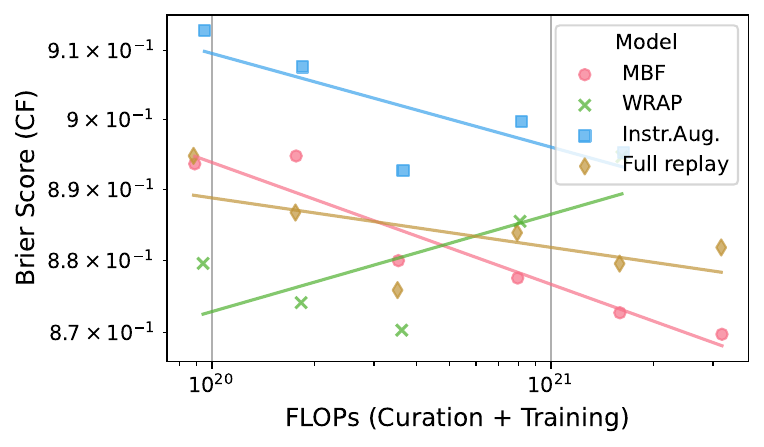}
        \caption{Medical MMLU CF tasks}
        \label{fig:a}
    \end{subfigure}
    \hfill
    \begin{subfigure}{0.49\linewidth}
        \centering
        \includegraphics[width=\linewidth]{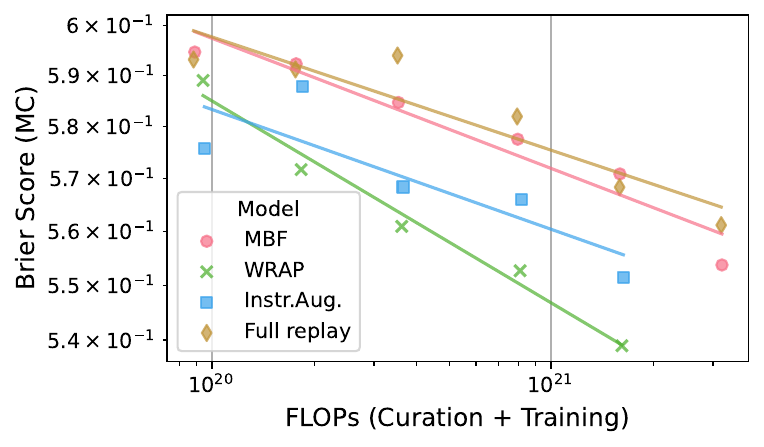}
        \caption{All medical MC tasks}
        \label{fig:med_scalling_curves_mc}
    \end{subfigure}
    \caption{Medical domain scaling curves on MMLU CF tasks and on MC tasks tasks. }
\end{figure*}

\begin{figure*}
    \centering
    \begin{subfigure}{0.49\linewidth}
        \centering
        \includegraphics[width=\linewidth]{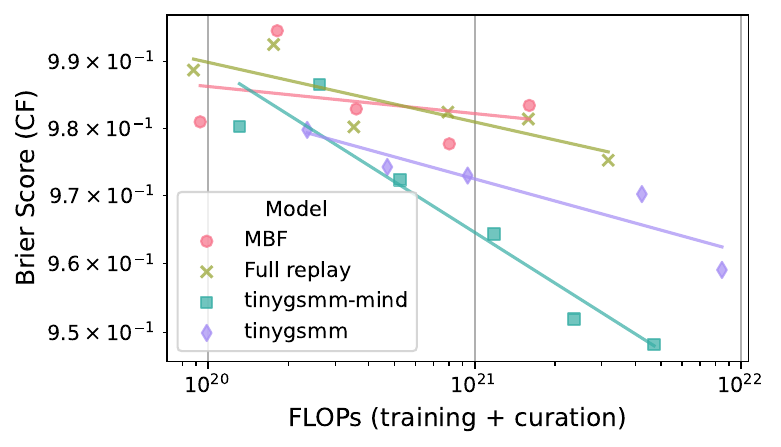}
        \caption{Medical MMLU CF tasks}
        \label{fig:math_scalling_curves_cf}
    \end{subfigure}
    \hfill
    \begin{subfigure}{0.49\linewidth}
        \centering
        \includegraphics[width=\linewidth]{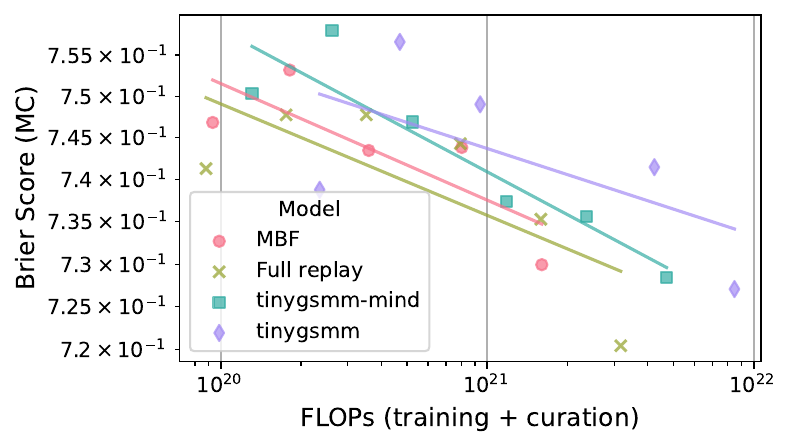}
        \caption{Medical MC tasks}
        \label{fig:math_scalling_curves_mc}
    \end{subfigure}
    \caption{Medical domain scaling curves on MMLU CF tasks and on MC tasks tasks. }
\end{figure*}

\begin{figure}[ht]
  \centering
    \includegraphics[width=0.5\linewidth]{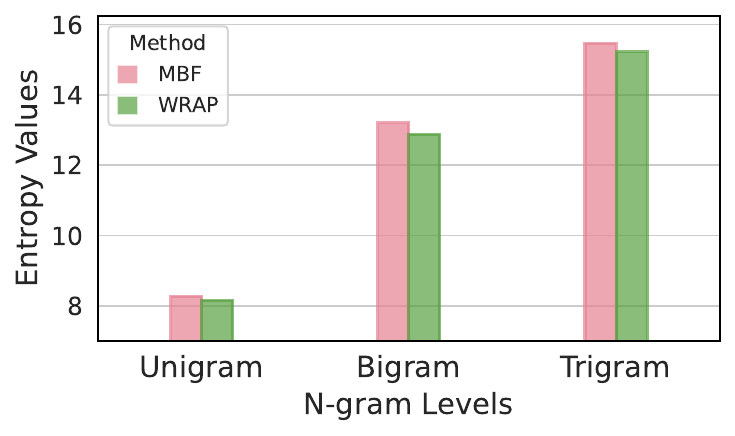}
    \caption{Comparison of entropy of the N-gram distribution; MBF exhibits higher entropy than WRAP which indicates greater diversity. As compute increases and the diversity of both \texttt{MBF} and \texttt{WRAP} increases. However, the diminishing performance of \texttt{WRAP} as compute increases, \texttt{MBF} offers more diverse document in the training data, while the number of unique documents in the \texttt{WRAP} dataset is lower for a fixed number of training tokens. Consequently, less diverse knowledge per unit of compute in \texttt{WRAP} leads to diminishing performance. To test this hypothesis, we measure corpus diversity in two ways. First, we compute the ratio of unique $n$-grams to total $n$-grams in the dataset, following \citet{li2015diversity}. Second, we calculate the entropy of the $n$-gram distribution, where higher entropy indicates greater diversity, reflecting a more uniform and less repetitive token distribution. Fig~\ref{fig:n-gram-diversity} presents the $n$-gram diversity scores for various values of $n$, while this Figure shows the corresponding entropy values. Both analyses confirm that \texttt{MBF} exhibits significantly higher diversity than \texttt{WRAP}, supporting our hypothesis.}
    \label{fig:entropy}
\end{figure}

\section{Cost calculation}

\label{sec:cost_estimation}

\begin{itemize}
    \item \( m \): Number of domain-specific ``seed'' tokens (e.g., obtained via MBF).
    \item \( k \): Expansion factor — number of synthetic tokens generated per seed token.
    \item \( a \): Per-token training cost, defined as \( a = 6\,|P| \), where \( |P| \) is the number of parameters of the training model.
    \item \( e \): Number of epochs over the upsampled tokens.
    \item \( |D| \): Effective total number of tokens the model sees during training.
    \item \( r \): Fraction of \( |D| \) that corresponds to the upsampled tokens.
    \item \( |C| = k\,m \): Total number of synthetic tokens generated.
    \item Relationship between upsampled data and total data size:
    \[
        r\,|D| = e(m + km) \quad \Rightarrow \quad |D| = \frac{e}{r} \bigl(m + k\,m\bigr)
    \]
    \item Seed token count as a function of dataset size:
    \[
        m = \frac{r\,|D|}{e(1 + k)}
    \]
    \item \textbf{Training Cost}:
    \[
        C_t = a\,|D|
    \]
    \item \textbf{Curation Cost}:
    \[
        C_g = c_s\,m + c_n\,k\,m
    \]
    \item \textbf{Total Cost} \( K \) is:
    \[
        K = C_t + C_g = a\,|D| + \left(c_s + k\,c_n\right)\,\frac{r\,|D|}{e(1 + k)}
    \]
    \item \( c_n \): Cost of generating a synthetic token, approximately \( 2 \times |P_g| \), where \( |P_g| \) is the number of parameters in the generation model.
    \item \( c_s \): Cost of obtaining seed tokens (e.g., via MBF), which includes both annotation and BERT model training cost.

   \begin{itemize}
        \item Assume \( m \) tokens are obtained from MBF by selecting documents with scores \( > 5 \). Then:
        \begin{align}
            c_s &= \frac{m\,R_{\mathrm{MBF},5}\,c_B + C_{\mathrm{BERT}}}{m} \\
                &= R_{\mathrm{MBF},5}\,c_B + \frac{C_{\mathrm{BERT}}}{m}
        \end{align}
        where:
        \begin{itemize}
            \item \( R_{\mathrm{MBF},5} \): MBF recall — number of tokens that need to be annotated to obtain \( m \) high-quality seed tokens (e.g., 22 for the medical domain in FineWebEdu).
            \item \( c_B \): Per-token inference cost of the BERT model.
            \item \( C_{\mathrm{BERT}} \): One-time cost of training the BERT annotator.
        \end{itemize}
    \end{itemize}
\end{itemize}

\subsection{Math Domain}
\label{app:math_cost}

To estimate the data curation cost for \textsc{TinyGSM}~\citep{liu2023tinygsm} and \textsc{TinyGSM-MIND}~\citep{olmo20242}, we make the following simplifying assumptions:

\begin{itemize}
    \item As before, we assume the inference cost per token is \( 2 \times |P| \), following~\citep{kaplan2020scaling}.
    \item \textsc{TinyGSM} uses GPT-3.5 to generate 12.3M synthetic math problems with Python solutions. Assuming GPT-3.5 has \( |P| = 175 \times 10^9 \) parameters (same as GPT-3).
    \item For simplicity, we omit the data filtering costs in both datasets.
\end{itemize}

\paragraph{Cost of \textsc{TinyGSM}.}
Given that \textsc{TinyGSM} consists of 1.8B tokens and the training cost is estimated at \( 350 \times 10^9 \) FLOPs/token, the total cost is:
\[
    K_{\textsc{TinyGSM}} = 1.8 \times 10^9 \times 350 \times 10^9 = 6.3 \times 10^{20} \text{ FLOPs}
\]

In annealing experiments, we use:
\begin{itemize}
    \item Batch size: 256
    \item Sequence length: 8192
    \item Upsampling ratio: 10\%
\end{itemize}

This results in \( 2.1 \times 10^6 \) tokens per step, of which \( 2.1 \times 10^5 \) are curated. The total compute cost for curation is:
\[
    K_{\textsc{TinyGSM}}(s) = s \times 2.1 \times 10^5 \times 350 \times 10^9
\]
where \( s \) is the number of annealing steps (e.g., 1k, 2k, 4k, 9k, 18k, 36k).

\paragraph{Cost of \textsc{TinyGSM-MIND}.}
\textsc{TinyGSM-MIND} rewrites the \textsc{TinyGSM} dataset using the 7B model \texttt{Qwen2.5-7B-Instruct}~\citep{yang2024qwen2}, resulting in 6.5B tokens — a \( 3.6\times \) upsampling ratio.

We estimate the curation cost as:
\[
    K_{\textsc{TinyGSM-MIND}}(s)
    = \frac{1}{3.6} \, K_{\textsc{TinyGSM}}(s)
    + \frac{2}{3.6} \, s \times 2.1 \times 10^5 \times 14 \times 10^9
\]
Here, \( 14 \times 10^9 \) is the assumed FLOPs per token for the 7B model.

\textit{Note:} Curation cost for \textsc{TinyGSM-MIND} is lower than for \textsc{TinyGSM} at the same number of steps because a larger portion of tokens (\( \frac{2}{3.6} \)) are curated using a smaller, more efficient model.

\end{document}